\pdfoutput=1
\def\year{2024}\relax
\documentclass[letterpaper]{article} 
\usepackage{aaai22}  
\usepackage{times}  
\usepackage{helvet}  
\usepackage{courier}  
\usepackage[hyphens]{url}  
\usepackage{graphicx} 
\usepackage{natbib}  
\usepackage{caption} 
\DeclareCaptionStyle{ruled}{labelfont=normalfont,labelsep=colon,strut=off} 
\usepackage{algorithm}
\usepackage{algorithmic}

\usepackage{enumitem}

\usepackage{newfloat}
\usepackage{listings}
\floatstyle{ruled}
\newfloat{listing}{tb}{lst}{}
\floatname{listing}{Listing}

\title{Comparing GPT-4 and Open-Source Language Models in Misinformation Mitigation}

\author{
    Tyler Vergho\textsuperscript{1}, 
    Jean-François Godbout\textsuperscript{2}, 
    Reihaneh Rabbany\textsuperscript{3}, 
    Kellin Pelrine\textsuperscript{3} \\
    \textsuperscript{1}Dartmouth College \quad
    \textsuperscript{2}Universit\'e de Montr\'eal \quad
    \textsuperscript{3}McGill University; Mila
}

\begin{document}
\maketitle
\begin{abstract}


Recent large language models (LLMs) have been shown to be effective for misinformation detection. However, the choice of LLMs for experiments varies widely, leading to uncertain conclusions. In particular, GPT-4 is known to be strong in this domain, but it is closed source, potentially expensive, and can show instability between different versions. Meanwhile, alternative LLMs have given mixed results. In this work, we show that Zephyr-7b presents a consistently viable alternative, overcoming key limitations of commonly used approaches like Llama-2 and GPT-3.5. This provides the research community with a solid open-source option and shows open-source models are gradually catching up on this task. We then highlight how GPT-3.5 exhibits unstable performance, such that this very widely used model could provide misleading results in misinformation detection. Finally, we validate new tools including approaches to structured output and the latest version of GPT-4 (Turbo), showing they do not compromise performance, thus unlocking them for future research and potentially enabling more complex pipelines for misinformation mitigation. 

\end{abstract}

\section{Introduction}
With the exponential growth of digital communications and increasingly sophisticated AI-driven misinformation \cite{Zhou2023}, determining the reliability and integrity of online content has become essential. As misinformation continues to pose a significant societal challenge, the role of advanced language models in detecting and addressing misinformation is increasingly critical. This paper examines the effectiveness of state-of-the-art models, particularly GPT-4, in misinformation detection and compares their performance with other emerging open-source models.

Specifically, this paper highlights an important development in the field: the emergence of smaller open-source models—in particular, finetuned versions of Mistral-7B, such as Zephyr alpha and beta---that begin to approach the performance of GPT-4. While GPT-4 continues to lead in certain tasks, the gap is narrowing, with open-source models like Zephyr achieving close to GPT-4 levels in detecting misinformation in the fake news datasets LIAR, LIAR-New, and CT-FAN-22. This trend is especially noteworthy as it suggests that smaller, more accessible models are beginning to offer a viable alternative to larger, proprietary systems in critical domains such as misinformation detection.

We also highlight how performance varies greatly between different models, and demonstrate that certain types of prompts can produce very poor results. Besides alternative open-source models that we found were not viable, we show that the performance of GPT-3.5 varies wildly. This is particularly noteworthy because it is likely the most widely used recent LLM in the field \cite{chen2023combating}. Our results suggest that there is a significant risk that the broad conclusions drawn from GPT-3.5 might be limited to intricacies of the specific prompt used and not easily generalizable to slightly different setups or different models.

Another notable contribution of this study is the use of structured JSON output on both proprietary (GPT-4) and open-source models, which represents to our knowledge the first time such an experiment is presented in the literature. Enforcing this structured output unlocks more systematic parsing and reliable analysis of the predictions made by these models in downstream applications. In our experiments, we particularly focused on the implementation of ``function calling" with GPT-4 and similar structured output formats on open-source models, using inference libraries that support imposing arbitrary grammars. With GPT-4, we observed that the performance using function calling is comparable to traditional methods. 


Our main contributions are:
\begin{itemize}[leftmargin=10pt,topsep=2pt,noitemsep]
    \item We show that unlike previous GPT-4 alternatives, Zephyr gives performance consistently matching or exceeding SLM and other pre gen-AI era SOTA across multiple datasets.
    \item We show that GPT-3.5 is very sensitive to details of the wording used in the prompts. This represents an important limitation since the conclusions reached by the model can be based on the input prompt, rather than reflecting the true performance of LLMs and more broadly applicable techniques.  
    \item We show how structured outputs can be obtained without performance penalty, making parsing more systematic and reliable and enabling more complex systems. We also show that GPT-4-1106 performance is similar to previous versions of GPT-4. These results provide the necessary validation to unlock new tools for research in this domain.
\end{itemize}

\section{Related Works}

In this section, we first discuss the widely-used OpenAI models for misinformation detection, which can give strong performance but also have several key limitations. We then discuss open-source alternatives.

\paragraph{OpenAI APIs for Misinformation Detection}

GPT-4 has emerged as a state-of-the-art model in misinformation detection and classification \cite{pelrine2023reliable, quelle2023perils}, demonstrating superior performance on multiple widely-used fake news datasets. Thus, combined with techniques to improve arbitrary LLM performance in this domain like HiSS \cite{zhang2023llmbased}, GPT-4 may have the strongest performance available. However, it also has several drawbacks. Its performance can be quite sensitive to the version used \cite{pelrine2023reliable, chen2023chatgpts}. And because it is a closed-source model, it can be expensive to use, and many of the tools or methods that require access to the model weights cannot be applied to improve its performance.

To avoid the cost limitation, many studies \cite{chen2023combating, zhang2023llmbased, 10174450, hu2023bad, Hoes2023} have instead focused on GPT-3.5, which is approximately 10x cheaper. But this too comes with a limitation: while advanced in many respects, it has yielded mixed results in misinformation detection. Some works have achieved state-of-the-art performance \cite{zhang2023llmbased} or found it matching or exceeding GPT-4 \cite{yu2023open}, while others find mixed results \cite{pelrine2023reliable, chen2023llmgenerated} or that it is outright worse than small language models \cite{hu2023bad}. Consequently, while an effective way to reduce costs in some cases, this model may be a risky alternative to GPT-4, potentially leading to erroneous conclusions about the efficacy of detecting  misinformation across different LLMs or methods. In our experiments, we confirm that this is indeed a real issue and demonstrate that certain open-source models may offer the flexibility and performance that closed source models like GPT-3.5 and GPT 4 lack.


\paragraph{Open-Source Models}

Open-source models solve several of the limitations of OpenAI and other closed models: they are more customizable, there is no risk of unexpected version updates or board revolts, and in many cases they can be much cheaper. However, performance of these models is a crucial concern. Previous studies have evaluated the performance of various large language models, such as Llama and Llama-2, in detecting misinformation. However, these studies have often yielded mixed results, indicating that not all models, especially open-source ones, are equally proficient in this domain \cite{yu2023open, chen2023large}. Recognizing this gap, our research aims to find an open-source model that can match or even surpass the performance benchmarks set by proprietary models like GPT-4 in misinformation detection datasets.

In selecting a suitable open-source model for our experiments, we chose to focus on Zephyr \cite{tunstall2023zephyr}, a fine-tuned version of the recently released Mistral model. Despite its relatively small size of 7 billion parameters, it has been shown to outperform 13B and even 70B-parameter models on certain benchmarks \cite{jiang2023mistral} as well as several other metrics on the Open LLM Leaderboard \cite{open-llm-leaderboard}. Its recent release ensures that the model has been exposed to contemporary data, which is crucial for the ever-evolving landscape of misinformation. Additionally, Zephyr's comparative performance against larger models, despite its smaller size, offers a unique opportunity to understand the efficiency and effectiveness of open-source models in misinformation detection. We also attempted experiments with instruction-finetuned versions of Llama-2 \cite{touvron2023llama}. However, these models consistently performed worse than the Mistral variants and significantly worse than GPT-4, a trend reflected by other studies as well \cite{yu2023open}.

Utilizing open-source models in misinformation detection is valuable for several reasons. Open-source models provide greater accessibility and transparency, allowing researchers and developers to scrutinize and trace the model's inner workings. In return, this can potentially enhance explainability approaches that require access to the original model weights \cite{Szczepaski2021}. This transparency---as well as enhanced privacy guarantees---is vital in trust-building, which is essential for misinformation detection applications \cite{mohseni2021machine}. Furthermore, open-source models can be adapted and improved by the broader community, potentially leading to greater advancements in this field. For instance, this paper directly leverages the capability to constrain generation and decoding on locally-run models to enforce structured JSON output, which simplifies processing in downstream applications \cite{geng2023grammarconstrained}. Additional techniques such as DoLa \cite{chuang2023dola} could potentially be applied to local models to reduce hallucinations and increase factuality in later research as well. 


\section{Methodology}


\paragraph{Open-Source Models}
For evaluating local models, particularly Zephyr, we employed a prompt that emulated the structured output format used in function calling with GPT-4. This prompt was carefully designed to match the template observed during the instruction fine-tuning phase of these models, ensuring consistency in the evaluation process. The exact prompt is available in the Appendix.


Unless otherwise specified, models that were run locally were quantized using the Q5\_KM quantization format. This method quantizes models to 5 bits, which saves memory and storage space while incurring minimal perplexity loss \cite{dettmers2023case}. Model quantization is an important consideration in practical applications where resource optimization is a priority.

\paragraph{Function Calling with GPT-4}

We prompt the LLM similarly to the Explain-then-Score prompt from \cite{pelrine2023reliable}, but provide a function for output using the OpenAI function calling API instead of generic unstructured chat. The exact prompting is detailed in the Appendix.

\paragraph{Baselines} We compare with several LLMs and prompting approaches from the literature. In particular, we consider Score Zero-Shot (``Score''), Binary Zero-Shot (``Binary''), and Explain-then-Score prompts from \cite{pelrine2023reliable}. This previous study provides results on two versions of GPT-4, and some results on GPT-3.5. We also draw from \cite{yu2023open}, which uses the Score prompt and provides more results on GPT-3.5 as well as results on Llama-2. Finally, we consider the zero-shot prompt from \cite{hu2023bad} (``Zero'') which provided results on GPT-3.5. This prompt is very similar to the Binary prompt---both ask for a binary output, and are nearly paraphrases of each other.

\paragraph{Data and Evaluation} We use standard datasets with input text and veracity labels: LIAR \cite{wang-2017-liar}, LIAR-New \cite{pelrine2023reliable}, and CT-FAN-22 (English) \cite{KohlerSSWS0S22}. The first two consist of short statements (a sentence or two), while the latter has significantly longer articles (often thousands of tokens) as input. For LIAR-New and CT-FAN (which have imbalanced classes) we report macro F1 score, while for LIAR we report accuracy. These metrics are the standard measures for evaluating performance on these datasets \cite{pelrine2023reliable}.\footnote{For CT-FAN, we follow the approach of \cite{pelrine2023reliable} to mark all predictions on the ill-defined ``other'' class as wrong, giving the most stringent evaluation.}

\section{Results}
\paragraph{Model Comparison}

\begin{table}[h]
\centering
\caption{Model performance on LIAR and LIAR-New. Results above the vertical line are taken from the literature, while those below come from our experiments.}
\resizebox{\linewidth}{!}{%
\begin{tabular}{l|c|c}
\hline
Model & LIAR-New & LIAR \\
\hline
GPT-4-0314 Explain-then-Score \cite{pelrine2023reliable} & 65.5 & 68.4 \\
GPT-4-0613 Explain-then-Score \cite{pelrine2023reliable} & 65.3 & 64.9 \\
GPT-4-0314 Score \cite{pelrine2023reliable} & 60.4 & 64.9 \\
GPT-3.5-0301 Score \cite{pelrine2023reliable} & 61.1 & 67.3 \\
GPT-3.5-0613 Binary \cite{pelrine2023reliable}& 55.7 & 53.6 \\
Llama-2-13B \cite{yu2023open} & - & 50.0 \\
Llama-2-70B \cite{yu2023open} & - & 49.1 \\
\hline
GPT-4-0314 Zero & 67.9 & 67.6 \\
GPT-4-0613 Zero & 66.7 & 65.1 \\
GPT-3.5-0301 Binary & 63.2 & 65.5 \\
GPT-3.5-0613 Score & 58.6 & 62.2 \\
GPT-3.5-0301 Zero & 54.5 & 57.3 \\
GPT-3.5-0613 Zero & 44.1 & 53.7 \\
zephyr-7b-alpha Explain-Then-Score & 62.2 & 62.8 \\
zephyr-7b-beta Explain-Then-Score & 63.2 & 62.9 \\
zephyr-7b-beta Zero & 61.4 & 58.8 \\
Mistral-7B-Instruct-v0.1 (f16) Explain-Then-Score & 59.2 & 57.1 \\
OpenOrca-Platypus2-13B Explain-Then-Score & 55.5 & 55.7 \\
\hline
\end{tabular}
}
\label{tab:liar-liar-new-model-performance}
\end{table}

The results from Table~\ref{tab:liar-liar-new-model-performance} indicate that Zephyr is a viable open-source model in this domain. It not only comes within a couple points of the GPT-4 Explain-then-Score results, but also surpasses GPT-4 Score Zero-Shot on LIAR-New. Other open-source models are producing poor results---e.g., Llama-2's performance here is almost random---so Zephyr provides a solid, validated option for contexts where an open-source LLM is important. It is also important to note that Zephyr outperforms much larger models, such as the 10 times bigger 70B version of Llama-2.


On LIAR, GPT-4 models demonstrate a slightly larger advantage over Zephyr, particularly with the Explain-then-Score construction. This divergence may partly stem from Zephyr's more recent training data, which does not provide as significant a benefit in this context as on the LIAR-New dataset (as the LIAR-New dataset was constructed to comprise data beyond GPT-4's knowledge cutoff date, but is not beyond Zephyr's). Meanwhile, GPT-3.5 gives unstable results. In some cases it gives excellent performance, such as the 0301 version on LIAR with Score prompt. In others, however, its performance is terrible, such as all cases with the Zero prompt. While GPT-4's performance varies, it isn't nearly to the same degree, and Zephyr too does not suffer such a performance drop with the Zero prompt. Consequently, usage of GPT-3.5 demands caution, such as using prompts it is known to work with or thorough testing of different prompts.

\begin{table}[h]
\centering
\caption{Model performance on CT-FAN. Zephyr zero-shot results are on par or better than the strongest known non-OpenAI models. While not as strong as the OpenAI ones, this shows that this open-source alternative is still viable with longer inputs.}
\resizebox{\linewidth}{!}{%
\begin{tabular}{l|c|c}
\hline
Model & English & German \\
\hline
GPT-4-0314 Score Zero-Shot \cite{pelrine2023reliable} & 42.8 & 38.7 \\
GPT-3.5-0301 Score Zero-Shot \cite{yu2023open} & 43.7 & - \\
Prev. English SOTA \cite{taboubi2022icompass} & 33.9 & -  \\
Prev. German SOTA \cite{TranK22-4} & - & 29.0  \\
RoBERTa-Large \cite{pelrine2023reliable} & 26.8 & - \\
Llama-2-13B \cite{yu2023open} & 21.2 & - \\
Llama-2-70B \cite{yu2023open} & 25.4 & - \\
\hline
zephyr-7b-beta Explain-Then-Score & 33.0 & 30.0 \\
\hline
\end{tabular}
}
\label{tab:ct-fan-model-performance}
\end{table}

In Table~\ref{tab:ct-fan-model-performance}, we examine the performance of several models on the CT-FAN dataset. Here, the Zephyr-7b-beta model achieves comparable performance to non-OpenAI SOTA, in both English and German. Since these older approaches are fine-tuned on CT-FAN, while Zephyr is not, this shows Zephyr is producing reasonable results, which again is not the case with alternatives like Llama-2. That said, this performance is notably lower than its relative performance on the LIAR and LIAR-New datasets. 
The decline in performance on CT-FAN could be attributed to several factors, including the complexity and diversity of the dataset, as well as the much larger prompt size in comparison to LIAR and LIAR-New. CT-FAN, with its nuanced and context-heavy content, might be testing the limits of Zephyr's capabilities, underscoring the challenges open-source models face in handling more complex misinformation scenarios. This contrast with GPT-4-0314's performance suggests that while open-source models like Zephyr are closing the gap in certain areas, they still lag in more complex misinformation detection tasks. Nonetheless, since Zephyr-7b does achieve comparable or better performance than the previous non-OpenAI, trained SOTA models, we conclude it is still a viable foundation model for work with this type of data.

\paragraph{GPT-4 Function Calling}

\begin{table}[h]
\centering
\small
\caption{GPT-4 Function Calling Comparison. It does not hurt performance, and for unknown reasons improves it dramatically with 1106 on CT-FAN.}
\resizebox{\linewidth}{!}{%
\begin{tabular}{l|c|c|c}
\hline
Model & LIAR-New & LIAR & CT-FAN (English) \\
\hline
GPT-4-0314 Explain-then-Score \cite{pelrine2023reliable} & 65.5 & 68.4 & 43.4 \\
GPT-4-0613 Explain-then-Score \cite{pelrine2023reliable} & 65.3 & 64.9 & 42.5 \\
GPT-4-0613 Function Calling & 64.0 & 65.5 & 42.3 \\
GPT-4-1106 Function Calling & 67.7 & 63.3 & 49.6 \\
\hline
\end{tabular}
}
\label{tab:gpt4-func-calling-comparison}
\end{table}

We next extend our structured prompting approach to GPT-4 through function calling, aiming to understand what impact this might have on performance. Our results, as shown in Table 2, reveal that function calling with GPT-4 effectively maintains high performance levels, approaching or exceeding the outcomes of the Explain-then-Score method. This demonstrates the feasibility of obtaining structured output without significantly compromising the model's accuracy in misinformation detection.


\paragraph{GPT-4 Updates}
\vspace{-4mm}
\begin{table}[h]
\centering
\small
\caption{Comparing GPT-4-1106 with GPT-4-0314 with Score prompt (latter from \cite{pelrine2023reliable}) The new version of GPT-4 has solid performance, and is improved on more recent data (LIAR-New).}
\resizebox{\linewidth}{!}{%
\begin{tabular}{l|c|c|c|c}
\hline
Model & LIAR-New & LIAR & CT-FAN English & CT-FAN German \\
\hline
GPT-4-0314 & 60.5 & 64.9 & 42.8 & 38.7 \\
GPT-4-1106-preview & 64.9 & 62.4 & 43.1 & 35.7 \\
\hline
\end{tabular}
}
\label{tab:gpt-turbo}
\end{table}

The results from Table 3, showcasing the performance of the latest GPT-4 iteration, GPT-4-1106, reveal that it holds up well against its predecessor which had the strongest performance in \cite{pelrine2023reliable}, GPT-4-0314, across various misinformation detection benchmarks. Notably, GPT-4-1106 demonstrates significantly improved performance on the LIAR-New dataset with an F1 score of 64.9, potentially reflecting its training on more recent data (the LIAR-New dataset is based on political statements made after the main knowledge cutoff of older versions of GPT-4, so many examples may reflect more recent events). However, on the LIAR dataset, it slightly lags behind GPT-4-0314, which scores 64.9 compared to 62.4 for GPT-4-1106. 

Since \cite{pelrine2023reliable, chen2023chatgpts} showed that GPT-4 performance could drop significantly between versions, this validation is critical before using the newest version in this domain. Our findings show that while the most recent version of GPT-4 does not significantly underperform, it also does not markedly outperform its predecessors, despite the more recent training data. On the one hand, this result opens up usage of the new model and its features, such as cheaper cost and the Assistants API. On the other, it suggests opportunities for smaller, open-source models to bridge the gap because GPT-4 is not significantly progressing in this domain. As these models continue to evolve and improve, they may soon rival the capabilities of proprietary models like GPT-4, especially in specific contexts like misinformation detection.

\section{Discussion}

In summary, this paper reveals a nuanced landscape where both proprietary and open-source models demonstrate varied misinformation detection capabilities, with some models distinctly outperforming others. Not all models, irrespective of being open-source or proprietary, demonstrate equal proficiency in misinformation detection. This is evident in the underwhelming and inconsistent performance of models like GPT-3.5 and Llama-2 on commonly applied misinformation detection benchmarks, underscoring the need for careful selection and evaluation of models in this domain.

A pivotal insight from this study is the emergence of smaller open-source models, specifically the versions of Zephyr, which begin to approach the performance of GPT-4 in this context. This highlights a significant shift towards the democratization of advanced AI technologies. The advantages of open-source models, such as accessibility, transparency, and adaptability, are particularly salient in the context of misinformation detection---a domain where trust and verifiability are paramount.

Additionally, the structured JSON output approach, applied to both GPT-4 and open-source models, has proven a useful methodology in the systematic parsing and reliable analysis of misinformation predictions with explanations attached. Past research has shown this ``Explain-then-Score'' approach improves performance beyond a simple binary classification baseline \cite{pelrine2023reliable}. This technique is particularly important when applied to open-source models, where unreliable instruction-following may detract from otherwise solid performance on misinformation detection benchmarks. Further research may expand on this approach in applications where a broader range of outputs may be necessary, such as when incorporating web retrieval to leverage additional context and evidence.

However, it is critical to recognize that not all models are equally effective. The variation in their performance levels, as demonstrated in our experiments, underscores the need to target ongoing research efforts in this area. The promising results of models like Zephyr indicate that certain open-source models are emerging as strong contenders on misinformation detection benchmarks, while commonly used alternatives like Llama-2 may not prove as effective for further research. Likewise, GPT-3.5-turbo is perhaps the most commonly used LLM in this domain \cite{chen2023combating}, but our results show that it is very sensitive to prompting, and thus could lead to uncertain conclusions. For example, contradictory results in the literature like \cite{pelrine2023reliable} (finding LLMs outperformed SLMs) and \cite{hu2023bad} (finding the reverse) might be due to the particular prompting and usage of the surprisingly unstable GPT-3.5.

\section{Conclusion}

In conclusion, this study extends past work on misinformation detection using open and proprietary advanced language models. While proprietary models like GPT-4 continue to lead in certain aspects, the evolving capabilities of open-source models offer new possibilities and avenues for research. The main takeaway is the identification of an open-source model, Zephyr, that demonstrates competitive performance in misinformation detection. This presents a promising tool for future research, contributing to the broader efforts to combat misinformation in an increasingly complex information landscape. We also showed risks associated with the most commonly used LLM for misinformation detection, GPT-3.5. Finally, we validated new tools including structured output and the latest version of GPT-4. 

Confirmation that readily available open-source models achieve comparable performance to the state-of-the-art on detecting ``fake news'' will hopefully contribute to the implementation of more reliable and pervasive real-world misinformation mitigation systems. In future work, we plan to investigate other recent open-source models such as Qwen \cite{qwen}. We also note that our investigation was limited to zero-shot approaches. Few-shot and especially finetuning approaches would be worthwhile areas for future research.

\section{Acknowledgements}

This work was partially funded by the CIFAR AI Chairs Program and by Berkeley SPAR. We thank SPAR for connecting collaborators to begin the project. 

\section{Author Contributions}

Tyler Vergho led the research, experiments, and writing for this project. Jean-Fran\c{c}ois Godbout and Reihaneh Rabbany advised the project, contributing ideas and feedback. Kellin Pelrine supervised the project, providing guidance and feedback at all stages.








\bibliography{ref}

\begin{thebibliography}{26}
\providecommand{\natexlab}[1]{#1}

\bibitem[{Bai et~al.(2023)Bai, Bai, Chu, Cui, Dang, Deng, Fan, Ge, Han, Huang, Hui, Ji, Li, Lin, Lin, Liu, Liu, Lu, Lu, Ma, Men, Ren, Ren, Tan, Tan, Tu, Wang, Wang, Wang, Wu, Xu, Xu, Yang, Yang, Yang, Yang, Yao, Yu, Yuan, Yuan, Zhang, Zhang, Zhang, Zhang, Zhou, Zhou, Zhou, and Zhu}]{qwen}
Bai, J.; Bai, S.; Chu, Y.; Cui, Z.; Dang, K.; Deng, X.; Fan, Y.; Ge, W.; Han, Y.; Huang, F.; Hui, B.; Ji, L.; Li, M.; Lin, J.; Lin, R.; Liu, D.; Liu, G.; Lu, C.; Lu, K.; Ma, J.; Men, R.; Ren, X.; Ren, X.; Tan, C.; Tan, S.; Tu, J.; Wang, P.; Wang, S.; Wang, W.; Wu, S.; Xu, B.; Xu, J.; Yang, A.; Yang, H.; Yang, J.; Yang, S.; Yao, Y.; Yu, B.; Yuan, H.; Yuan, Z.; Zhang, J.; Zhang, X.; Zhang, Y.; Zhang, Z.; Zhou, C.; Zhou, J.; Zhou, X.; and Zhu, T. 2023.
\newblock Qwen Technical Report.
\newblock \emph{arXiv preprint arXiv:2309.16609}.

\bibitem[{Beeching et~al.(2023)Beeching, Fourrier, Habib, Han, Lambert, Rajani, Sanseviero, Tunstall, and Wolf}]{open-llm-leaderboard}
Beeching, E.; Fourrier, C.; Habib, N.; Han, S.; Lambert, N.; Rajani, N.; Sanseviero, O.; Tunstall, L.; and Wolf, T. 2023.
\newblock Open LLM Leaderboard.
\newblock \url{https://huggingface.co/spaces/HuggingFaceH4/open_llm_leaderboard}.

\bibitem[{Caramancion(2023)}]{10174450}
Caramancion, K.~M. 2023.
\newblock Harnessing the Power of ChatGPT to Decimate Mis/Disinformation: Using ChatGPT for Fake News Detection.
\newblock In \emph{2023 IEEE World AI IoT Congress (AIIoT)}, 0042--0046.

\bibitem[{Chen and Shu(2023{\natexlab{a}})}]{chen2023llmgenerated}
Chen, C.; and Shu, K. 2023{\natexlab{a}}.
\newblock Can LLM-Generated Misinformation Be Detected?
\newblock arXiv:2309.13788.

\bibitem[{Chen and Shu(2023{\natexlab{b}})}]{chen2023combating}
Chen, C.; and Shu, K. 2023{\natexlab{b}}.
\newblock Combating Misinformation in the Age of LLMs: Opportunities and Challenges.
\newblock \emph{arXiv preprint arXiv:2311.05656}.

\bibitem[{Chen, Zaharia, and Zou(2023)}]{chen2023chatgpts}
Chen, L.; Zaharia, M.; and Zou, J. 2023.
\newblock How is ChatGPT's behavior changing over time?
\newblock arXiv:2307.09009.

\bibitem[{Chen et~al.(2023)Chen, Wei, Cao, Zhou, and Hu}]{chen2023large}
Chen, M.; Wei, L.; Cao, H.; Zhou, W.; and Hu, S. 2023.
\newblock Can Large Language Models Understand Content and Propagation for Misinformation Detection: An Empirical Study.
\newblock arXiv:2311.12699.

\bibitem[{Chuang et~al.(2023)Chuang, Xie, Luo, Kim, Glass, and He}]{chuang2023dola}
Chuang, Y.-S.; Xie, Y.; Luo, H.; Kim, Y.; Glass, J.; and He, P. 2023.
\newblock DoLa: Decoding by Contrasting Layers Improves Factuality in Large Language Models.
\newblock arXiv:2309.03883.

\bibitem[{Dettmers and Zettlemoyer(2023)}]{dettmers2023case}
Dettmers, T.; and Zettlemoyer, L. 2023.
\newblock The case for 4-bit precision: k-bit Inference Scaling Laws.
\newblock arXiv:2212.09720.

\bibitem[{Geng et~al.(2023)Geng, Josifoski, Peyrard, and West}]{geng2023grammarconstrained}
Geng, S.; Josifoski, M.; Peyrard, M.; and West, R. 2023.
\newblock Grammar-Constrained Decoding for Structured NLP Tasks without Finetuning.
\newblock arXiv:2305.13971.

\bibitem[{Hoes, Altay, and Bermeo(2023)}]{Hoes2023}
Hoes, E.; Altay, S.; and Bermeo, J. 2023.
\newblock Leveraging ChatGPT for Efficient Fact-Checking.

\bibitem[{Hu et~al.(2023)Hu, Sheng, Cao, Shi, Li, Wang, and Qi}]{hu2023bad}
Hu, B.; Sheng, Q.; Cao, J.; Shi, Y.; Li, Y.; Wang, D.; and Qi, P. 2023.
\newblock Bad Actor, Good Advisor: Exploring the Role of Large Language Models in Fake News Detection.
\newblock arXiv:2309.12247.

\bibitem[{Jiang et~al.(2023)Jiang, Sablayrolles, Mensch, Bamford, Chaplot, de~las Casas, Bressand, Lengyel, Lample, Saulnier, Lavaud, Lachaux, Stock, Scao, Lavril, Wang, Lacroix, and Sayed}]{jiang2023mistral}
Jiang, A.~Q.; Sablayrolles, A.; Mensch, A.; Bamford, C.; Chaplot, D.~S.; de~las Casas, D.; Bressand, F.; Lengyel, G.; Lample, G.; Saulnier, L.; Lavaud, L.~R.; Lachaux, M.-A.; Stock, P.; Scao, T.~L.; Lavril, T.; Wang, T.; Lacroix, T.; and Sayed, W.~E. 2023.
\newblock Mistral 7B.
\newblock arXiv:2310.06825.

\bibitem[{Köhler et~al.(2022)Köhler, Shahi, Struß, Wiegand, Siegel, 0001, and Schütz}]{KohlerSSWS0S22}
Köhler, J.; Shahi, G.~K.; Struß, J.~M.; Wiegand, M.; Siegel, M.; 0001, T.~M.; and Schütz, M. 2022.
\newblock Overview of the CLEF-2022 CheckThat! Lab: Task 3 on Fake News Detection.
\newblock In Faggioli, G.; 0001, N.~F.; Hanbury, A.; and Potthast, M., eds., \emph{Proceedings of the Working Notes of CLEF 2022 - Conference and Labs of the Evaluation Forum, Bologna, Italy, September 5th - to - 8th, 2022}, volume 3180 of \emph{CEUR Workshop Proceedings}, 404--421. CEUR-WS.org.

\bibitem[{Mohseni et~al.(2021)Mohseni, Yang, Pentyala, Du, Liu, Lupfer, Hu, Ji, and Ragan}]{mohseni2021machine}
Mohseni, S.; Yang, F.; Pentyala, S.; Du, M.; Liu, Y.; Lupfer, N.; Hu, X.; Ji, S.; and Ragan, E. 2021.
\newblock Machine learning explanations to prevent overtrust in fake news detection.
\newblock In \emph{Proceedings of the international AAAI conference on web and social media}, volume~15, 421--431.

\bibitem[{Pelrine et~al.(2023)Pelrine, Imouza, Thibault, Reksoprodjo, Gupta, Christoph, Godbout, and Rabbany}]{pelrine2023reliable}
Pelrine, K.; Imouza, A.; Thibault, C.; Reksoprodjo, M.; Gupta, C.; Christoph, J.; Godbout, J.-F.; and Rabbany, R. 2023.
\newblock Towards Reliable Misinformation Mitigation: Generalization, Uncertainty, and GPT-4.
\newblock arXiv:2305.14928.

\bibitem[{Quelle and Bovet(2023)}]{quelle2023perils}
Quelle, D.; and Bovet, A. 2023.
\newblock The Perils \& Promises of Fact-checking with Large Language Models.
\newblock arXiv:2310.13549.

\bibitem[{Szczepański et~al.(2021)Szczepański, Pawlicki, Kozik, and Choraś}]{Szczepaski2021}
Szczepański, M.; Pawlicki, M.; Kozik, R.; and Choraś, M. 2021.
\newblock New explainability method for BERT-based model in fake news detection.
\newblock \emph{Scientific Reports}, 11(1).

\bibitem[{Taboubi, Nessir, and Haddad(2022)}]{taboubi2022icompass}
Taboubi, B.; Nessir, M. A.~B.; and Haddad, H. 2022.
\newblock iCompass at CheckThat! 2022: combining deep language models for fake news detection.
\newblock \emph{Working Notes of CLEF}.

\bibitem[{Touvron et~al.(2023)Touvron, Martin, Stone, Albert, Almahairi, Babaei, Bashlykov, Batra, Bhargava, Bhosale, Bikel, Blecher, Ferrer, Chen, Cucurull, Esiobu, Fernandes, Fu, Fu, Fuller, Gao, Goswami, Goyal, Hartshorn, Hosseini, Hou, Inan, Kardas, Kerkez, Khabsa, Kloumann, Korenev, Koura, Lachaux, Lavril, Lee, Liskovich, Lu, Mao, Martinet, Mihaylov, Mishra, Molybog, Nie, Poulton, Reizenstein, Rungta, Saladi, Schelten, Silva, Smith, Subramanian, Tan, Tang, Taylor, Williams, Kuan, Xu, Yan, Zarov, Zhang, Fan, Kambadur, Narang, Rodriguez, Stojnic, Edunov, and Scialom}]{touvron2023llama}
Touvron, H.; Martin, L.; Stone, K.; Albert, P.; Almahairi, A.; Babaei, Y.; Bashlykov, N.; Batra, S.; Bhargava, P.; Bhosale, S.; Bikel, D.; Blecher, L.; Ferrer, C.~C.; Chen, M.; Cucurull, G.; Esiobu, D.; Fernandes, J.; Fu, J.; Fu, W.; Fuller, B.; Gao, C.; Goswami, V.; Goyal, N.; Hartshorn, A.; Hosseini, S.; Hou, R.; Inan, H.; Kardas, M.; Kerkez, V.; Khabsa, M.; Kloumann, I.; Korenev, A.; Koura, P.~S.; Lachaux, M.-A.; Lavril, T.; Lee, J.; Liskovich, D.; Lu, Y.; Mao, Y.; Martinet, X.; Mihaylov, T.; Mishra, P.; Molybog, I.; Nie, Y.; Poulton, A.; Reizenstein, J.; Rungta, R.; Saladi, K.; Schelten, A.; Silva, R.; Smith, E.~M.; Subramanian, R.; Tan, X.~E.; Tang, B.; Taylor, R.; Williams, A.; Kuan, J.~X.; Xu, P.; Yan, Z.; Zarov, I.; Zhang, Y.; Fan, A.; Kambadur, M.; Narang, S.; Rodriguez, A.; Stojnic, R.; Edunov, S.; and Scialom, T. 2023.
\newblock Llama 2: Open Foundation and Fine-Tuned Chat Models.
\newblock arXiv:2307.09288.

\bibitem[{Tran and Kruschwitz(2022)}]{TranK22-4}
Tran, H.-N.; and Kruschwitz, U. 2022.
\newblock ur-iw-hnt at CheckThat!-2022: Cross-lingual Text Summarization for Fake News Detection.
\newblock In Faggioli, G.; 0001, N.~F.; Hanbury, A.; and Potthast, M., eds., \emph{Proceedings of the Working Notes of CLEF 2022 - Conference and Labs of the Evaluation Forum, Bologna, Italy, September 5th - to - 8th, 2022}, volume 3180 of \emph{CEUR Workshop Proceedings}, 740--748. CEUR-WS.org.

\bibitem[{Tunstall et~al.(2023)Tunstall, Beeching, Lambert, Rajani, Rasul, Belkada, Huang, von Werra, Fourrier, Habib, Sarrazin, Sanseviero, Rush, and Wolf}]{tunstall2023zephyr}
Tunstall, L.; Beeching, E.; Lambert, N.; Rajani, N.; Rasul, K.; Belkada, Y.; Huang, S.; von Werra, L.; Fourrier, C.; Habib, N.; Sarrazin, N.; Sanseviero, O.; Rush, A.~M.; and Wolf, T. 2023.
\newblock Zephyr: Direct Distillation of LM Alignment.
\newblock arXiv:2310.16944.

\bibitem[{Wang(2017)}]{wang-2017-liar}
Wang, W.~Y. 2017.
\newblock {``}Liar, Liar Pants on Fire{''}: A New Benchmark Dataset for Fake News Detection.
\newblock In Barzilay, R.; and Kan, M.-Y., eds., \emph{Proceedings of the 55th Annual Meeting of the Association for Computational Linguistics (Volume 2: Short Papers)}, 422--426. Vancouver, Canada: Association for Computational Linguistics.

\bibitem[{Yu et~al.(2023)Yu, Yang, Pelrine, Godbout, and Rabbany}]{yu2023open}
Yu, H.; Yang, Z.; Pelrine, K.; Godbout, J.~F.; and Rabbany, R. 2023.
\newblock Open, Closed, or Small Language Models for Text Classification?
\newblock arXiv:2308.10092.

\bibitem[{Zhang and Gao(2023)}]{zhang2023llmbased}
Zhang, X.; and Gao, W. 2023.
\newblock Towards LLM-based Fact Verification on News Claims with a Hierarchical Step-by-Step Prompting Method.
\newblock arXiv:2310.00305.

\bibitem[{Zhou et~al.(2023)Zhou, Zhang, Luo, Parker, and De~Choudhury}]{Zhou2023}
Zhou, J.; Zhang, Y.; Luo, Q.; Parker, A.~G.; and De~Choudhury, M. 2023.
\newblock Synthetic Lies: Understanding AI-Generated Misinformation and Evaluating Algorithmic and Human Solutions.
\newblock In \emph{Proceedings of the 2023 CHI Conference on Human Factors in Computing Systems}, CHI ’23. ACM.

\end{thebibliography}

\clearpage
\section{Appendix: Detailed Prompts}

\vspace{0.25cm}

\subsection{Function Calling with GPT-4}

Function calling is a feature exposed by OpenAI API. This method allows users to define specific tasks or `functions' for the model to execute. Function calling is akin to calling a function in traditional programming, where a set of instructions is executed to return a specific output. In the context of misinformation detection, function calling is particularly valuable as it allows for structured and consistent output, a critical requirement for the accurate analysis of information and its eventual processing by downstream applications.

For our experiments, we used a structured prompt that was designed to guide GPT-4 in evaluating the truthfulness of statements and providing a corresponding score and explanation. This approach was intended to test the model's ability to discern and articulate the accuracy of information in a clear and structured manner. The prompt, shown below, mirrors the phrasing used in prior studies \cite{pelrine2023reliable}, but has been adapted for the function calling context.

\label{appendix-fc}
\begin{lstlisting}
{
  "name": "detect_misinformation",
  "description": "Rate the truthfulness of a statement",
  "parameters": {
    "type": "object",
    "properties": {
      "score": {
        "type": "number",
        "description": "A score from 0 to 100, where 0 represents definitively false and 100 represents definitively true, based on the available context."
      },
      "explanation": {
        "type": "string",
        "description": "An explanation for evaluation of the statement and the final score. Do not calculate the score until you've presented a thorough analysis."
      }
    },
    "required": ["score", "explanation"]
  }
}

messages=[{
    "role": "user", 
    "content": "Rate the truthfulness of the following statement: {statement}"
}]
\end{lstlisting}

\subsection{Local Model Evaluation}

To achieve a structured and parseable output format, we utilized the llama.cpp library, which allows for the specification of a grammar in Backus–Naur form. This feature was instrumental in enforcing the output to be in valid JSON format. Enforcing JSON output is conceptually similar to the function calling approach with GPT-4, aiming to streamline the parsing and analysis process.

Where possible, we construct prompts for instruction-tuned open-source models based on the prompt template used during finetuning. Here, we base the following prompt on the model card for Zephyr that is published on the HuggingFace website.

\begin{lstlisting}
<|system|>
Your task is to provide a score from 0 to 100, where 0 represents definitively false and 100 represents definitively true, but you must not state your score until you've presented a thorough analysis. Return a JSON object with two keys. First, your analysis in the \"explanation\" key, then a score with the \"score\" key. The score should match the analysis and your assessment of the statement's truthfulness. You must always provide a score.
</s>
<|user|>
Rate the truthfulness of the following statement: {statement}
</s>
<|assistant|>
\end{lstlisting}

\vspace{0.25cm}

For CT-FAN and Zephyr-beta, we switch to the following prompt, which we observe provides more consistent performance across larger prompts.

\vspace{0.25cm}

\begin{lstlisting}
<|user|>
Rate the truthfulness of the following statement: {statement}</s>
<|user|>
Your task is to provide a score from 0 to 100, where 0 represents definitively false and 100 represents definitively true, but you must not state your score until you've presented a thorough analysis. Return a JSON object with two keys. First, your analysis in the \"explanation\" key, then a score with the \"score\" key. The score should match the analysis and your assessment of the statement's truthfulness. You must always provide a score.</s>
<|assistant|>
\end{lstlisting}

\end{document}